%% file: main.tex
\documentclass[letterpaper, 10 pt, conference]{ieeeconf}  

\IEEEoverridecommandlockouts                              

\usepackage{amsmath,amssymb,amsfonts}
\usepackage{algorithmic}
\usepackage{graphicx}
\usepackage{textcomp}
\usepackage{xcolor}
\usepackage{tabularx}
\usepackage{makecell}
\usepackage{multirow}
\usepackage{hyperref}
\usepackage{mathrsfs} 
\usepackage{balance} 

\usepackage[capitalize]{cleveref}
\crefname{section}{Sec.}{Secs.}
\Crefname{section}{Section}{Sections}
\Crefname{table}{Table}{Tables}
\crefname{table}{Tab.}{Tabs.}
\crefname{figure}{fig.}{figs.} 
\Crefname{figure}{Fig.}{Figs.} 
\usepackage{subfiles} 

\usepackage{xurl} 

\title{\LARGE \bf
LiDAR-MIMO: Efficient Uncertainty Estimation\\
for LiDAR-based 3D Object Detection
}

\author{
Matthew Pitropov$^{1}$,
Chengjie Huang$^{2}$,
Vahdat Abdelzad$^{1}$,
Krzysztof Czarnecki$^{1}$,
and Steven Waslander$^{3}$
\thanks{$^{1}$Department of Electrical and Computer Engineering, University of Waterloo, Canada. Contacts: \href{mailto:matthew.pitropov@uwaterloo.ca}{matthew.pitropov@uwaterloo.ca}, \href{mailto:vahdat.abdelzad@uwaterloo.ca}{vahdat.abdelzad@uwaterloo.ca}, \href{mailto:krzysztof.czarnecki@uwaterloo.ca}{krzysztof.czarnecki@uwaterloo.ca}}
\thanks{$^{2}$David R. Cheriton School of Computer Science, University of Waterloo, Canada. Contact: \href{mailto:c.huang@uwaterloo.ca}{c.huang@uwaterloo.ca}}
\thanks{$^{3}$University of Toronto Robotics Institute, Canada. Contact: \href{mailto:steven.waslander@utoronto.ca}{steven.waslander@utoronto.ca}}
}

\def\etal{\emph{et al}.}

\begin{document}

\maketitle
\thispagestyle{empty}
\pagestyle{empty}

\begin{abstract}
\subfile{sections/abstract}
\end{abstract}

\subfile{sections/body}

\section*{ACKNOWLEDGMENT} 
\subfile{sections/acknowledgement}


\balance 
\bibliographystyle{IEEEtran}
\bibliography{IEEEabrv, citations}

\end{document}

%% file: sections/abstract.tex
The estimation of uncertainty in robotic vision, such as 3D object detection, is an essential component in developing safe autonomous systems aware of their own performance. However, the deployment of current uncertainty estimation methods in 3D object detection remains challenging due to timing and computational constraints. To tackle this issue, we propose LiDAR-MIMO, an adaptation of the multi-input multi-output (MIMO) uncertainty estimation method to the LiDAR-based 3D object detection task. Our method modifies the original MIMO by performing multi-input at the feature level to ensure the detection, uncertainty estimation, and runtime performance benefits are retained despite the limited capacity of the underlying detector and the large computational costs of point cloud processing. We compare LiDAR-MIMO with MC dropout and ensembles as baselines and show comparable uncertainty estimation results with only a small number of output heads. Further, LiDAR-MIMO can be configured to be twice as fast as MC dropout and ensembles, while achieving higher mAP than MC dropout and approaching that of ensembles.

%% file: sections/body.tex
\section{INTRODUCTION}
Deep neural networks (DNNs) are currently indispensable for implementing perception in robotics, including autonomous vehicles. Since perception is inherently uncertain, real-time and reliable estimation of perceptual uncertainty is essential for the safe operation of such systems. Object detectors based on DNNs produce a confidence score, e.g., max softmax, for each detection. However, such scores are inaccurate estimates of the actual probability of perception errors, that is, predictive uncertainty~\cite{hendrycks17baseline}.

Bayesian Neural Networks (BNNs)~\cite{neal1996bayesian} are a disciplined approach to provide accurate estimates of uncertainty. The key idea of BNNs is to place a prior probability distribution over the network parameters and use Bayesian inference to determine a posterior. Bayesian inference is computationally expensive and intractable  for the large network sizes used in autonomous driving perception tasks such as object detection. Thus, much of the current work has focused on approximations that provide uncertainty estimates with a reasonable computation cost. Deep ensembles~\cite{lakshminarayanan2017simple} and Monte Carlo (MC) dropout~\cite{gal2016dropout} are well established approximation methods and have also been applied to object detection~\cite{feng2021review,miller2019benchmarking}. The key idea is to sample multiple models from an approximation of the posterior distribution and compute uncertainty as a statistic over their outputs (e.g., variance of the outputs). The need to perform multiple network passes makes these methods computationally expensive and thus challenging to deploy in constrained real-time environments such as autonomous vehicles.

\begin{figure}[t]
\hspace*{-0.5cm}
\centering
\includegraphics[width=1.0\columnwidth]{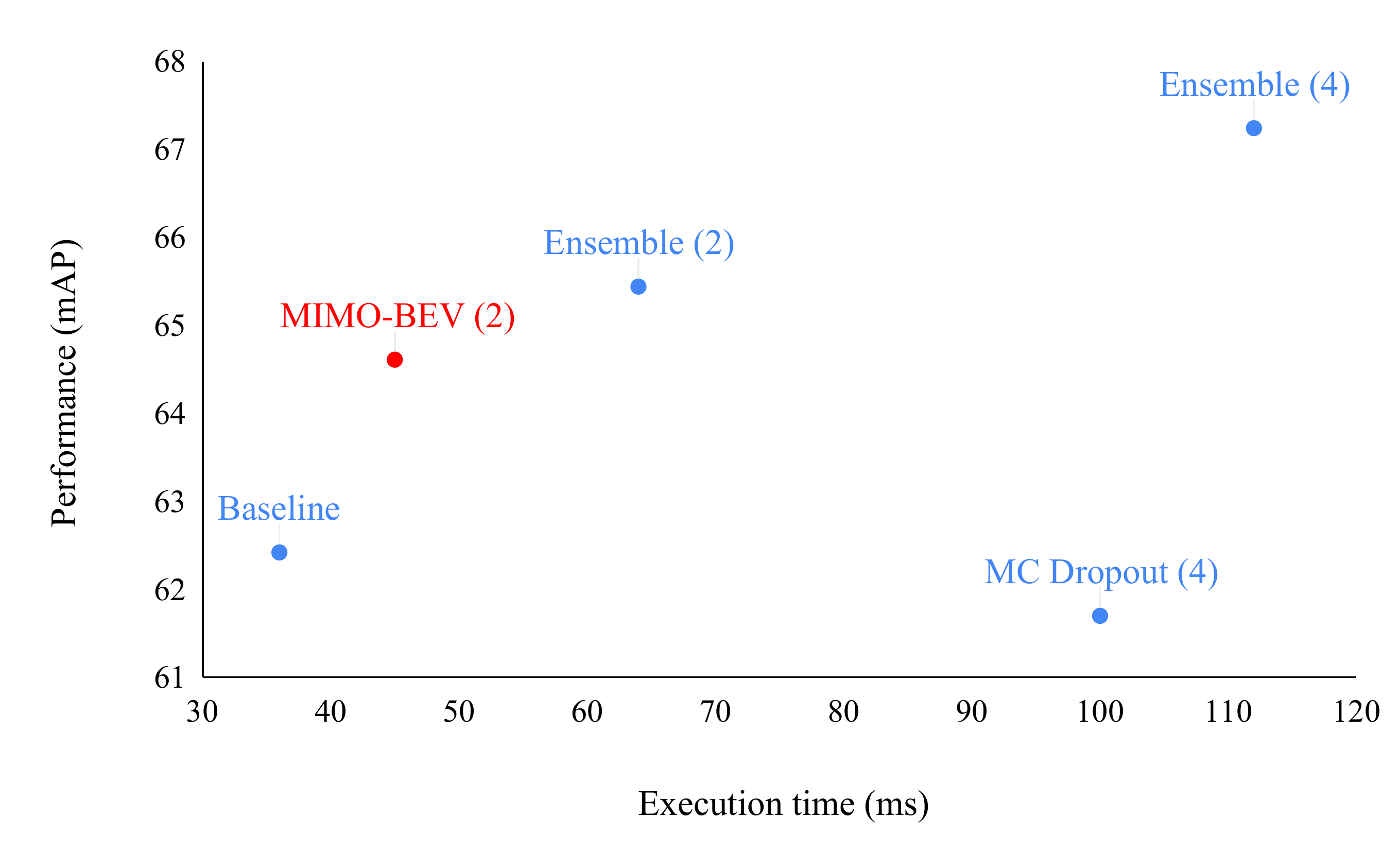}
\caption{The detection performance (mAP) compared to the inference time (execution time) for different uncertainty estimation methods (including baseline and proposed one in red) of the PointPillars model trained on the KITTI dataset. The number inside parentheses for the ensemble indicates the number of instances used for uncertainty estimation.}
\label{f:map_runtime}
\end{figure}

To address this challenge, we propose LiDAR-MIMO, an adaptation of the multi-input multi-output (MIMO) uncertainty estimation method~\cite{havasi2021training} to the task of 3D object detection in LiDAR point clouds. MIMO is a method shown to provide state-of-the-art uncertainty estimates in image classification, but at a fraction of the computational cost of the existing methods. The key idea behind the MIMO method is to train a single network with multiple inputs and outputs (one per input) and then feed a single image in multiple times during inference, utilizing the resulting output distribution to estimate predictive uncertainty. 

Although images are easily combined by stacking in the MIMO method, adapting MIMO to 3D LiDAR object detection is not as straightforward. It requires combining multiple input point clouds, which can be costly and potentially negate the performance advantage of MIMO. Therefore, we propose to extend the original idea such that the multi-input happens at the feature level. It eliminates having the voxelization bottleneck and transfer such a bottleneck to the middle layers of the network. We perform multi-input over the bird’s-eye-view (BEV) feature maps. This is different from the original MIMO which stacks raw inputs, but is key for the LiDAR-based 3D object detection. We associate the success of this extension with the fact that the voxel feature network does not have enough capacity to have a significant contribution to uncertainty estimation.

We evaluate the resulting LiDAR-MIMO design using two different 3D object detectors and compare them with MC dropout and ensembles. We demonstrate that LiDAR-MIMO has comparable performance with MC dropout and ensemble in terms of uncertainty estimation quality, as measured by scoring rules and calibration errors. However, it achieves this while being twice as fast as MC dropout and ensemble, and having a higher detection performance (mAP) than both the vanilla DNN (baseline) and MC dropout, as shown in \Cref{f:map_runtime}. The baseline model is trained without any extensions for uncertainty estimation. \par
The main contributions of this paper are as follows: (i) We are the first to adapt the MIMO architecture to LiDAR-based 3D object detection, confirming the effectiveness of this method in this new task. (ii) We modify the MIMO architecture to combine inputs at feature level rather than raw inputs to reduce data processing, which is key for LiDAR point clouds in real-time constrained environments. Further, this modification also takes better advantage of the limited available capacity in the underlying object detector compared to the original approach. (iii) We extensively evaluate LiDAR-MIMO over multiple 3D object detectors and show LiDAR-MIMO has higher mAP than MC dropout, comparable performance on uncertainty assessments, and is only marginally slower than the underlying vanilla detector. As a result, LiDAR-MIMO can be can be twice as fast as using MC dropout or ensembles.

\section{BACKGROUND AND RELATED WORKS}

\subsection{3D object detection for LiDAR point clouds}

LiDAR-based 3D object detection has emerged as a pivotal component of autonomous driving systems, with point-based~\cite{qi2018frustum,shi2019pointrcnn}, voxel-based~\cite{ku2018joint, yan2018second, zhou2018voxelnet, lang2019pointpillars}, and hybrid approaches showing steadily improvement~\cite{shi2020pv}. Voxel-based approaches that exploit the bird's-eye view (BEV) projection, such as  PointPillars (PP) \cite{lang2019pointpillars} and SECOND (SC) \cite{yan2018second}, achieve competitive detection performance at low computational cost, which makes them attractive for deployment on AVs. Information for BEV feature map creation is listed within the papers. Although hybrid approaches, like PV-RCNN~\cite{shi2020pv}, have higher mean average precision (mAP), they also add substantial computational cost. For example, PV-RCNN~\cite{shi2020pv} achieves $82.97$\% mAP on the KITTI benchmark, compared to $80.29$\% mAP for SC, but SC runs 2x faster than PV-RCNN \cite{qian20213d}.

\begin{figure*}[ht]
\centering
\includegraphics[width=1.0\textwidth]{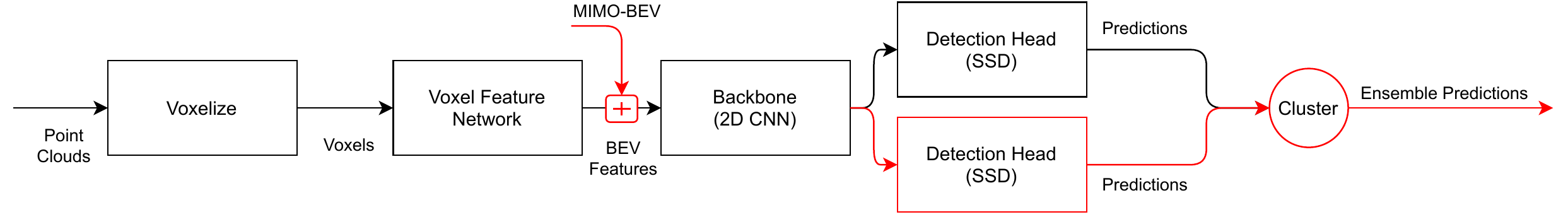}
\caption{The network architecture of a LiDAR-MIMO 3D object detector with additions highlighted in red}
\label{f:mimo_diagram}
\end{figure*}

\Cref{f:mimo_diagram} highlights the main components of a voxel-based detector using PP and SC as examples (ignore the elements in red for now). The input point cloud is first voxelized into pillar (PP) or cuboid (SC) voxels. These are fed into a voxel feature extraction (VFE) network, which is PointNet~\cite{qi2017pointnet} for PP and sparse 3D convolutions for SC. The BEV projection of the resulting features, or BEV feature maps, are then passed to a backbone, which applies 2D convolutions. Finally, the backbone features are passed to a detection head, which outputs classes and bounding boxes of the detected objects. The BEV feature maps are sometimes referred to as a ``2D pseudo image''~\cite{qi2017pointnet}.

\subsection{Uncertainty estimation for classification}
The predictive uncertainty of a classification network has two components~\cite{NIPS2017_7141}: epistemic uncertainty, which is due to the uncertainty over the model parameters caused by the limited availability of data; and aleatoric uncertainty, which is due to the inherent noise in the data itself. The predictive distribution output by a network should adequately reflect both sources of uncertainty, which is the case for Bayesian inference for NNs. However, since exact such inference is computationally intractable, practical realizations of BNNs that place a distribution on their weights are necessarily approximations, relying on making prior assumptions about the class of distributions for the weights and using different approximate inference methods, including variational Bayes~\cite{blundell2015weight} and Markov Chain Monte Carlo methods~\cite{neal1996bayesian}. These methods are highly sensitive to hyper-parameters and difficult to scale to large datasets and network architectures~\cite{maddox2019simple}.

A more practical class of approximate BNNs, which we discuss in the remainder of this section, rely on some form of model ensembling that samples models from the approximate posterior and combines their outputs to compute the resulting predictive distribution. For classification, the predictive distribution is simply the mean of the categorical distributions from each ensemble member~\cite{Gal2016Uncertainty}. For regression, each member is usually designed as a variance network ~\cite{nix1994estimating}, which is trained with two outputs, the mean and variance, using the so-called aleatoric regression loss~\cite{NIPS2017_7141}. The overall predictive distribution is then characterized by the mean and variance of the resulting mixture distribution~\cite{lakshminarayanan2017simple}.
We now briefly introduce ensembles and MC dropout, which are the two main practical methods for approximating BNNs found in literature, and MIMO, which aims at reducing the computational cost of these methods. \par
\par
\noindent \textbf{Deep ensembles~\cite{lakshminarayanan2017simple}:}
A state of the art baseline in uncertainty estimation is to train multiple models with the same architecture and ensure their diversity by random initialization of weights and training data shuffling. Ensembling is shown to outperform MC dropout on uncertainty prediction metrics~\cite{lakshminarayanan2017simple}. In particular, it shows much stronger diversity of outputs than MC dropout, since each model is trained independently~\cite{havasi2021training}. Although easily parallelized, deep ensembles remain very costly in terms of both processing and memory use.

\noindent \textbf{MC dropout~\cite{gal2016dropout}:}
MC dropout employs dropout both during training and inference and obtains multiple predictions by performing multiple forward passes with different dropout network realizations for the same input. Normally dropout layers are added throughout the network as a regularization method to reduce over-fitting~\cite{srivastava2014dropout}.  Interestingly, MC dropout has been shown to be equivalent to approximate Bayesian inference in deep Gaussian processes~\cite{gal2016dropout}.

\noindent \textbf{MIMO~\cite{havasi2021training}:}
Similar to MC dropout, a multi-input multi-output (MIMO) network represents an ensemble as a single network, but unlike MC dropout, it requires only a single forward pass during inference. The main idea is to stack multiple inputs as a combined input into a single network with multiple copies of its classification head, one per input. The network is trained by stacking different inputs and then giving each classification head the ground truth label of the corresponding input for its loss function. During inference, instead of stacking different inputs, the same input is stacked to satisfy the input tensor size requirement. Each head makes its own prediction for the same input, which results in the desired sample set of outputs. MIMO relies on the fact that modern DNNs are overparameterized and uses the available network capacity to fit multiple independent subnetworks. Because each output head is trained by independently sampled inputs, given sufficient capacity, the MIMO predictions can match the diversity of a deep ensemble~\cite{havasi2021training}.

When compared with MC dropout and ensembles, the MIMO method was demonstrated to provide a prediction time similar to a vanilla deterministic network. MIMO achieved this while outperforming the classification accuracy of MC dropout, and approaching that of an ensemble~\cite{havasi2021training}. It also had Negative Log-Likelihood (NLL) and Expected Calibration Error (ECE) \cite{naeini2015obtaining} scores comparable to deep ensembles, indicating uncertainty estimation of similarly high quality. Recently, MIMO has been adapted for the task of 2D object detection showing similar performance results ~\cite{cygert2021robust}.

\subsection{Uncertainty estimation for object detection}

Uncertainty estimation has recently also attracted interest in the context of object detection.
For this task, uncertainty estimation methods are used both for object classification and bounding box-regression, which results in bounding boxes being represented by probability distributions. In contrast to classification, object detection also needs to deal with multiple object hypotheses, and thus steps such as non-maximum suppression (NMS) need to be refined or extended to deal with bounding boxes as probability distributions.

Most closely related to our work, Feng \etal~\cite{8569814} propose a probabilistic LiDAR-based 3D vehicle detector with dropout layers and a variance regression output for predicted variances. They find that adding the variance regression output improves vehicle detection accuracy. The authors did not evaluate uncertainty estimation, but the results show that the occlusion level and vehicle distance correlate with the aleatoric uncertainty, and the vehicle detection accuracy correlates with the epistemic uncertainty.

Similarly, Zhong \etal~\cite{zhong2020uncertaintyaware} modify the SECOND architecture with a variance regression output and capture aleatoric uncertainty via an adapted aleatoric loss function. The loss improves the uncertainty prediction for angular parameters by using the von Mises distribution. 

Harakeh and Waslander~\cite{harakeh2021estimating} conduct an in-depth survey and evaluation of predictive uncertainty for deep 2D object detection architectures using different aleatoric regression losses, MC dropout, and ensemble methods. To evaluate the methods, they use mAP, calibration error, and scoring rules. The results highlight i) the importance of scoring rules for capturing the reliability of output distributions, ii) issues with NMS while seeking output sample sets, and iii) computational and performance limitations of ensemble and MC dropout methods. Our work follows in this vein of rigorous evaluation of predictive uncertainty, and addresses the high computational cost of MC dropout and ensembles.

We are unaware of any work adapting MIMO for uncertainty estimation for 3D object detection or LiDAR inputs.

\section{LIDAR-MIMO 3D OBJECT DETECTION}
\Cref{f:mimo_diagram} depicts our proposed adaptation of the MIMO method to LiDAR-based 3D object detection, with the components of the underlying 3D object detector (e.g, PointPillars or SECOND) in black and the extensions required by LiDAR-MIMO in red. We propose a design variant for combining multiple inputs named MIMO-BEV. In MIMO-BEV, we stack BEV feature maps of point cloud inputs as multiple inputs. In the following sections, we describe the input combination, the loss functions, and the clustering of the predictions from the multiple heads.

\subsection{Input combination}
In order to provide our 3D object detector DNN with multiple inputs we went through three potential variants. We named these variants MIMO-noID, MIMO-ID and MIMO-BEV. 

The MIMO-noID variant combined the point clouds from different frames resulting in one large point cloud. This resulted in low mAP performance as the detection heads could not focus on features from a single input. To fix this, we decided to create the MIMO-ID variant. This variant, adds an additional head ID feature to each point to allow the network to differentiate among the inputs.

\begin{figure*}[ht!]
\centering
\includegraphics[width=1.0\textwidth]{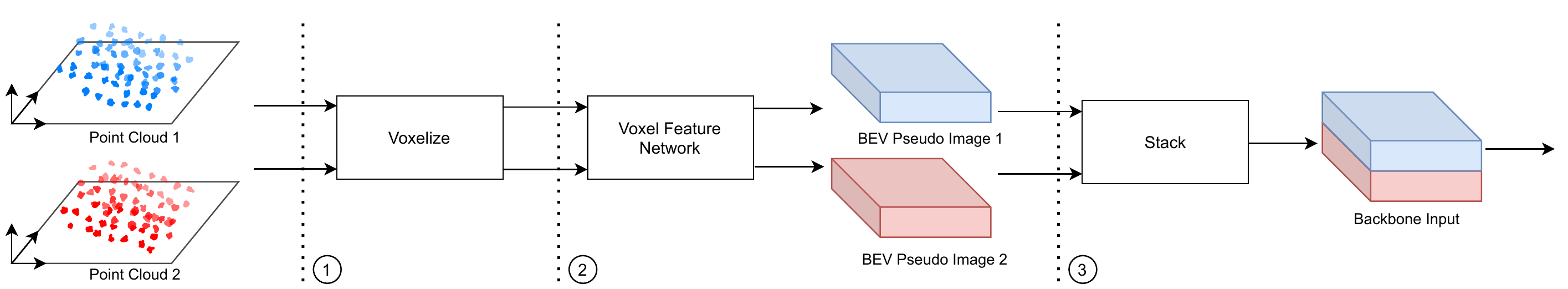}
\caption{Pseudo image stacking for MIMO-BEV with two inputs}
\label{f:mimo_bev_training}
\end{figure*}

For the MIMO-BEV variant, the main idea is that the equivalent of stacking images for MIMO in image classification is stacking the BEV pseudo image in 3D object detection. \Cref{f:mimo_bev_training} depicts the MIMO-BEV architecture for training. In step 1, one point cloud for each detection head is individually voxelized, followed by step 2 where those voxels are sent into the VFE network. Note that the voxelization and VFE functions take one input point cloud and are called separately for each input. The input point clouds can be from different frames or the same frame, with the likelihood depending on the chosen MIMO parameters. These parameters are the number of heads, input repetition, and batch repetition and are discussed in Sec.~4. At this point each point cloud has been converted into separate BEV psuedo images (feature maps), which are then stacked together in step 3 as input for the Backbone.  During testing, we instead use the following modified steps to reduce execution time while keeping the same performance. A single point cloud is voxelized in step 1 and sent through the VFE network for step 2. The output, which is the pseudo image, is duplicated multiple times depending on the number of heads in step 3. Then, these the pseudo images are stacked in step 4 and used as input to the backbone.

In contrast to MIMO-BEV, MIMO-ID can also fit subnetworks into the VFE network, rather than just the Backbone. MIMO-ID showed two issues, however. The need to duplicate and modify the point cloud input for each MIMO head increased data processing time by a factor 2.8, which increased the total inference time by a factor of 1.7 compared to MIMO-BEV for PointPillars. Also, while the variant produces similar uncertainty estimation results as MIMO-BEV for PointPillars, it does not for SECOND, possibly because of capacity issues in the VFE of the latter. Due to page limitations and performance we do not include the numerical results for MIMO-ID.

\subsection{Loss functions}
\noindent \textbf{Classification loss:}
We use a softmax version of the original focal loss \cite{lin2017focal}, which allows us to use uncertainty evaluation metrics such as the Brier Score that require a probability distribution over all classes. This is different from the default choice in PointPillars and SECOND implementations, which use one sigmoid per class.

\noindent \textbf{Regression on linear parameters:}
To obtain variances for the bounding box regression values, we add an extra convolution layer as output to the detection head(s) for heteroscedastic regression~\cite{NIPS2017_7141}. For such a regression, the noise is data-dependent and must be learned as part of the loss function, see Equation \eqref{eqn:reg_var_loss} \cite{8569814, NIPS2017_7141}. The ground-truth bounding box $\mathbf{y}_{g t}$ and predicted bounding box $\mathbf{u_x}$ parameters include seven regression variables: centroid position, $x, y, z$, length, width, height and orientation. The log variance, $\lambda= \log{\sigma^2}$, for each parameter is regressed for numerical stability, to avoid the possibility of dividing by zero \cite{NIPS2017_7141}. When data has high uncertainty, the first term becomes 0 and the model's loss only contains the second term. This causes the loss to be less affected by data with higher uncertainty.

\begin{equation}
\label{eqn:reg_var_loss}
L_{v a r}=
\frac{1}{2} \exp (-\lambda^T)
(\mathbf{y}_{gt}-\mathbf{u}_{\mathbf{x}})^{2}
+\frac{1}{2}\lambda^T\boldsymbol{1}
\end{equation}

\noindent \textbf{Regression on angular parameters:}
For regression on linear parameters,  the loss function is derived from the Negative Log-Likelihood (NLL) of the Gaussian density function. In contrast, for regression on angular parameters, which are periodic, a periodic approximation of the Gaussian distribution should be used~\cite{zhong2020uncertaintyaware}. Thus, we use Equation
\eqref{eqn:reg_var_ang_loss}, derived from the NLL of the von Mises PDF (see \cite{zhong2020uncertaintyaware} for details).

\begin{equation}
\label{eqn:reg_var_ang_loss}
\begin{gathered}
L_{var_{\theta}}=\log I_{0}(\exp (-\lambda))-\exp (-\lambda) \cos \left(\theta-\theta_{t}\right) \\
+ \lambda_{V}\textit{ELU}\left(\lambda-\lambda_{0}\right)
\end{gathered}
\end{equation}

\subsection{Output clustering and merging}
In the MIMO architecture, it is necessary to merge outputs to obtain the final predictions (see~\Cref{f:mimo_diagram}). This is straightforward for image classification, because of the single object per input to be classified. In contrast, for object detection, we may have multiple objects per input and thus need to cluster and merge the detections of the same object from multiple heads. One might consider applying NMS to the combined output of all models, however, this does not take into account the number of objects, models or object classes. To group predictions, we adopted the consensus clustering method, shown to perform best compared to the affirmative (at least one) and unanimous (all members must agree)  methods~\cite{casado2020ensemble}. Consensus clustering requires more than the number of outputs divided by 2 in order for the cluster to be valid. To merge predictions, we used the mean method. Roza \etal~\cite{roza2020assessing} implemented several different box merging strategies, and showed significant variations on sample-based uncertainty estimates resulting from different strategies. Based on their results, we used the mean method, which had the best performance overall.
Orientation angles need a special consideration, however, since they could differ by 180{\textdegree} in the outputs.
To handle this case, we instead take the angle from the box with the highest confidence to remove the negative effect of mean methods on bimodal distributions. For example, two clustered predictions with 0{\textdegree} and 180{\textdegree} would result in worse prediction with a mean angle of 90\textdegree.

\section{EXPERIMENTS}
In this section, we evaluate our proposed LiDAR-MIMO architecture against other uncertainty estimation methods on publicly available AV datasets. We demonstrate that LiDAR-MIMO produces on par uncertainty estimates while having lower inference latency, which is paramount for safety-critical and resource-constrained applications such as AVs.

\subsection{Datasets}
All experiments were performed through training with the KITTI dataset~\cite{Geiger2012CVPR}. The KITTI dataset is a widely-used multimodal dataset for autonomous driving in clear weather. The full training set of the dataset is used to train each model. For mAP and partition counts, we report the results the on the full validation (val) set. For uncertainty evaluation, we follow Feng \etal~\cite{feng2019can} and split the val set equally into the recalibration (recal) set and the evaluation (eval) set. The recal set is used to determine score thresholds and calibrate the models. The eval set is then used to test the calibrated model by calculating the scoring rules and calibration errors.

\subsection{Models}

We evaluate LiDAR-MIMO against MC dropout and ensembles on two detectors, PointPillars (PP) and SECOND (SC). We use the PP and SC implementations provided in OpenPCDet~\cite{openpcdet2020}, and extend them with MC dropout, ensembling, and MIMO-BEV. We refer to the resulting detectors for MC dropout, ensemble and MIMO-BEV as \emph{multi-output}. More efficient ensemble approaches such as TreeNet~\cite{lee2015m} or BatchEnsemble~\cite{wen2020batchensemble} were not tested due to their lower performance compared to a standard ensemble~\cite{havasi2021training}. 

\noindent \textbf{Baseline:} The baseline models for evaluating detection performance use the original PP and SC implementations. For KITTI, we use the PP and SC models provided with OpenPCDet. All baseline models are trained with the sigmoid focal loss for 80 epochs.

\noindent \textbf{Multi-output:} Each multi-input model shares the same voxel-feature network, backbone, and head from the original PP and SC, with some exceptions for MC dropout and MIMO-BEV as described shortly. Further, all multi-output detectors use the softmax focal classification loss, smooth L1 regression losses, and the previously described aleatoric regression losses, and share the same clustering and merging stage. Each multi-input model is trained for an extra 40 epochs compared to the baseline models in order to have the models converge with the new loss functions.

\noindent \textbf{MC dropout:} We create a variant of the backbone by inserting a dropout layer, with 0.5 probability, after the ReLU of each deconvolution block. During inference, we keep the dropout layers active for MC dropout. This model is trained with a batch size of 6 for 120 epochs.

\noindent \textbf{Ensemble:} We train four unique models to form our ensemble. Each model is trained with a different random seed for frame order shuffling, for a batch size of 6 with 120 epochs.

\noindent \textbf{MIMO-BEV:} MIMO-BEV requires additional code for psuedo images  combination and adding multiple heads. There are three hyper parameters for MIMO: number of heads, input repetition (IR) and batch repetition (BR). We selected the number of heads to be 2 as the base networks have low capacity. IR is used to select a percentage of frame groupings in a batch to have matching frame, while BR duplicates each frame in a batch for smoother training. Since we are using a low batch size we were able to train with an IR of 0\% and a BR of 0. This model is trained with a batch size of 3 with 120 epochs.

\cref{tab:kitti_pp_timing} specifies the runtime parameters for the evaluated models. The number of forward passes is the number of times that a model is run for input to output(s). The number of outputs is the number of NMS outputs after all forward passes are run. For example, an MC dropout model is run 4 times to get 4 outputs, and a MIMO model is run once for 2 outputs, since it is set for 2 heads.
We use two heads since our experiments have shown that PointPillars and SECOND do not have sufficient capacity to accommodate three or more independent subnetworks, and adding more capacity would come with increased runtime. This is consistent with the original MIMO work~\cite{havasi2021training}, where the optimal number of subnetworks for the studied image classification networks is 2-3. The MC dropout and ensemble output amounts have been selected to have comparable performance with our MIMO model while also having low execution time.

\begin{table*}[ht]
\footnotesize
\centering
\caption{The inference time (in \textit{ms}) for the PointPillars models trained on the KITTI dataset}
\label{tab:kitti_pp_timing}
\begin{tabular}{ c|ccccccc } 
 \Xhline{2\arrayrulewidth}
\multirow{2}{*}{Model} & \multirow{2}{*}{\begin{tabular}[c]{@{}c@{}}\# Fwd\\ Passes\end{tabular}} & \multirow{2}{*}{Outputs} & \multirow{2}{*}{\begin{tabular}[c]{@{}c@{}}Minimum\\ Cluster Size\end{tabular}} & \begin{tabular}[c]{@{}c@{}}Data\\ Processing\end{tabular} & VFE & \begin{tabular}[c]{@{}c@{}}Backbone\\ + Heads \end{tabular} & \multirow{2}{*}{\begin{tabular}[c]{@{}c@{}}Total\\ Time $\downarrow$\end{tabular}} \\ \cline{5-7}
 & & & & \multicolumn{3}{c}{\tiny{1 Fwd Pass}} &  \\
 \hline
 Baseline & 1 & 1 & - & 16 & 4 & 16 & 36 \\
 \hline
 MC dropout & 4 & 4 & 3 & 16 & 4 & 20 & 100 \\
 Ensemble (4) & 4 & 4 & 3 & 16 & 4 & 20 & 112 \\
 Ensemble (2) & 2 & 2 & 2 & 16 & 4 & 20 & 64 \\
 \hline
 MIMO-BEV & 1 & 2 & 2 & 16 & 4 & 25 & \textbf{45} \\
 \Xhline{2\arrayrulewidth}
\end{tabular}
\end{table*}

\subsection{Evaluation metrics}
We evaluate models based on the detection performance, inference time, and uncertainty estimates for both classification and regression.

\noindent \textbf{Detection:} 3D object detection algorithms are generally evaluated using the average precision (AP) metric, first introduced in the PASCAL VOC 2007 object detection challenge~\cite{everingham2010pascal}. The calculation of AP is based on the area under the precision-recall curve calculated for 40 recall points.

\noindent \textbf{Classification uncertainty:}
We evaluate this uncertainty based on the softmax distribution extracted for each prediction using two metrics: negative log-likelihood (NLL) and Brier score, which are both proper~\cite{gneiting2007strictly}. The NLL score is a local scoring rule which only evaluates the softmax distribution at the ground truth label. The Brier score \cite{brier1950verification} is a non-local and strictly proper scoring rule which evaluates the whole softmax distribution by taking the squared error between a predictive probability from the network and its one-hot encoded ground truth label. For our qualitative results we calculate the Shannon entropy (SE) along with the mutual information (MI)~\cite{Gal2016Uncertainty}. SE contains the total predictive uncertainty (epistemic and aleatoric) while MI captures only the epistemic uncertainty.

To evaluate the classification calibration, we use an extension of Average Calibration Error (ACE) called Marginal Calibration Error (MCE)~\cite{kumar2019verified}. ACE calculates the absolute error averaged over all score intervals equally and only considers the uncertainty calibration quality on the object's ground-truth class. In contrast, MCE measures the uncertainty of a classifier’s predicted distribution over all classes of interest.

\noindent \textbf{Regression uncertainty:}
For regression parameters, we evaluate uncertainty based on the predicted means and variances for each prediction that can be matched to a ground-truth object. We use NLL and energy scores~\cite{gneiting2008assessing} as proper scoring rules. The energy score is a strictly proper and non-local score, analogous to the Brier score for classification. It is derived from energy distance. To calculate it, we use an efficient Monte-Carlo approximation for multivariate Gaussians~\cite{gneiting2008assessing}. For our qualitative results we calculate the spatial uncertainty for a 3D bounding box using epistemic and aleatoric total variance (ETV and ATV)~\cite{8569814}. The total variance is the trace of the covariance matrix. For ETV, the covariance matrix is created using the regression values from a bounding box cluster while for ATV it is created using the predicted variances from a bounding box cluster.

For the calibration error, we use a method created by Kuleshov \etal~\cite{kuleshov2018accurate}, which is an extension to calibration methods for classification. Instead of the confidence scores being used to place predictions into bins based on thresholds, the CDF is used to place predictions.

\noindent \textbf{Detection partitions:}
We focus on three partitions in order to see clear differences in the uncertainty metrics~\cite{hoiem2012diagnosing, harakeh2021estimating}. True Positives (TP) are predictions that match a ground-truth (GT) box with an IoU over the TP IoU threshold. Mislocalized false positives (FP\textsubscript{ML}) are predictions that match with a GT box, but their IoU is below the TP IoU threshold and higher than 0.1. The last partition is background false positives (FP\textsubscript{BG}), where the prediction has a lower than 0.1 IoU with all GT boxes and thus it is highly likely that the prediction should be background.

\subsection{Results}

\subsubsection{The inference time}
\cref{tab:kitti_pp_timing} shows the inference time of each uncertainty estimation method for PointPillars and the KITTI dataset. All timings were calculated using a computer with an i7-8700K CPU and Titan Xp GPU. Data processing time is the time to load the point cloud and perform voxelization. For each model except MC dropout, the calculation for total time is data processing time added to the number of forward passes multiplied by the time for data to pass through the network. For the MC dropout model an efficient implementation can have the VFE network output cached and reused for each forward pass that encounters the dropout layers. Due to this caching, the calculation adds the data processing time to the VFE time, which is then added to the number of forward passes multiplied by the backbone and heads time.
As seen in \cref{tab:kitti_pp_timing}, MIMO-BEV is 60\% faster than ensemble and 55\% faster than the MC dropout.

\subsubsection{Average precision and detection partitions}
For each network architecture, we show separate tables with mAP and counts for individual detection partitions.

\cref{tab:kitti_pp_map} and \cref{tab:kitti_second_map} contain the results for their respective models trained on the KITTI dataset. In both cases, the MIMO-BEV model exceeds the mAP score of the MC dropout and baseline model, while approaching the ensemble models. The MIMO models have higher FP\textsubscript{BG} than the baseline model; however, their FP\textsubscript{ML} is lower than all others except ensemble. The ensemble models have the largest increase in mAP after clustering; for example, the individual member outputs scored $\sim$62-63\% mAP for PointPillars on KITTI. The ensemble also has the lowest FP\textsubscript{BG} count. \Cref{f:map_runtime} plots mAP and execution time for the PointPillars models trained on KITTI. The smaller improvement of MIMO over the SECOND baseline may indicate that the latter makes better use of its available capacity than PointPillars. 

\begin{table}[ht]
\footnotesize
\centering
\caption{The mAP and detection partitions for the PointPillars models trained over the KITTI dataset}
\label{tab:kitti_pp_map}
\begin{tabular}{ c|cccc }
 \Xhline{2\arrayrulewidth}
 \multirow{2}{*}{Model} & mAP & TP & FP\textsubscript{ML} & FP\textsubscript{BG} \\
 \cline{3-5}
  &  (\%) $\uparrow$ & \multicolumn{3}{c}{\tiny{Counts Per Partition}} \\
 \hline
  Baseline & 62.42 & 8128 & 4204 & 47028 \\
 \hline
  MC dropout & 61.70 & 7969 & 4600 & 39274 \\
  Ensemble (4) & \textbf{67.24} & 8406 & \textbf{3452} & \textbf{31844} \\
  Ensemble (2) & 65.44 & \textbf{8428} & 4099 & 82842 \\
  \hline
  MIMO-BEV & 64.61 & 8332 & 3857 & 56988 \\
\Xhline{2\arrayrulewidth}
\end{tabular}
\end{table}

\begin{table}[ht]
\footnotesize
\centering
\caption{The mAP and detect partitions for the SECOND models trained over the KITTI dataset}
\label{tab:kitti_second_map}
\begin{tabular}{ c|cccc }
 \Xhline{2\arrayrulewidth}
 \multirow{2}{*}{Model} & mAP & TP & FP\textsubscript{ML} & FP\textsubscript{BG} \\
 \cline{3-5}
  &  (\%) $\uparrow$ & \multicolumn{3}{c}{\tiny{Counts Per Partition}} \\
 \hline
  Baseline & 65.37 & 8397 & 3837 & 36713 \\
 \hline
  MC dropout & 64.89 & 8371 & 3833 & 33541 \\
  Ensemble (4) & \textbf{68.82} & 8654 & \textbf{3213} & \textbf{24871} \\
  Ensemble (2) & 67.77 & \textbf{8656} & 3702 & 60640 \\
  \hline
  MIMO-BEV & 65.45 & 8534 & 3581 & 44113 \\
 \Xhline{2\arrayrulewidth}
\end{tabular}
\end{table}

\subsubsection{Uncertainty evaluation metrics}
\begin{table*}[ht!]
\footnotesize
\centering
\caption{Uncertainty evaluation for the PointPillars models trained over the KITTI dataset}
\label{tab:kitti_pp_ue}
\begin{tabular}{ c|ccc|ccc|cc|cc|c|c }
 \Xhline{2\arrayrulewidth}
 \multirow{2}{*}{Model} & \multicolumn{3}{c|}{NLL (Cls) $\downarrow$} & \multicolumn{3}{c|}{Brier Score $\downarrow$} & \multicolumn{2}{c|}{NLL (Reg) $\downarrow$} & \multicolumn{2}{c|}{Energy Score $\downarrow$} & MCE & CE \\
 \cline{2-11}
  & TP & FP\textsubscript{ML} & FP\textsubscript{BG} & TP & FP\textsubscript{ML} & FP\textsubscript{BG} & TP & FP\textsubscript{ML} & TP & FP\textsubscript{ML} & (Cls) $\downarrow$ & (Reg) $\downarrow$ \\
 \hline
  MC dropout & \textbf{0.3521} & \textbf{0.4096} & 1.0870 & \textbf{0.1941} & \textbf{0.2359} & 0.8306 & -4.8418 & -2.1827 & 0.4540 & \textbf{0.6580} & 0.2296 & \textbf{0.0602} \\
  Ensemble (4) & 0.3569 & 0.4260 & 1.0130 & 0.1960 & 0.2454 & 0.7752 & -4.8780 & -2.2568 & 0.4458 & 0.6813 & \textbf{0.1995} & 0.0688 \\
  Ensemble (2) & 0.3558 & 0.4177 & 1.0362 & 0.1943 & 0.2406 & 0.7923 & -4.9023 & -2.2127 & 0.4422 & 0.6774 & 0.2160 & 0.0694 \\

  \hline
  MIMO-BEV & 0.3652 & 0.4445 & \textbf{0.9870} & 0.2011 & 0.2609 & \textbf{0.7633} & \textbf{-5.0020} & \textbf{-2.3396} & \textbf{0.4332} & 0.6704 & 0.2070 & 0.0642 \\
 \Xhline{2\arrayrulewidth}
\end{tabular}
\end{table*}

\begin{table*}[ht!]
\footnotesize
\centering
\caption{Uncertainty evaluation for the SECOND models trained over the KITTI dataset}
\label{tab:kitti_second_ue}
\begin{tabular}{ c|ccc|ccc|cc|cc|c|c }
 \Xhline{2\arrayrulewidth}
 \multirow{2}{*}{Model} & \multicolumn{3}{c|}{NLL (Cls) $\downarrow$} & \multicolumn{3}{c|}{Brier Score $\downarrow$} & \multicolumn{2}{c|}{NLL (Reg) $\downarrow$} & \multicolumn{2}{c|}{Energy Score $\downarrow$} & MCE & CE \\
 \cline{2-11}
  & TP & FP\textsubscript{ML} & FP\textsubscript{BG} & TP & FP\textsubscript{ML} & FP\textsubscript{BG} & TP & FP\textsubscript{ML} & TP & FP\textsubscript{ML} & (Cls) $\downarrow$ & (Reg) $\downarrow$ \\
 \hline
  MC dropout & \textbf{0.3211} & \textbf{0.372} & 1.1448 & \textbf{0.1663} & \textbf{0.2039} & 0.8853 & \textbf{-5.8268} & -1.4789 & \textbf{0.2835} & 0.5553 & 0.2352 & \textbf{0.0504} \\
  Ensemble (4) & 0.3521 & 0.3915 & 1.0957 & 0.1929 & 0.2182 & 0.8412 & -5.5626 & -2.3645 & 0.3151 & 0.5664 & 0.2078 & 0.0657 \\
  Ensemble (2) & 0.3509 & 0.3971 & 1.0840 & 0.1911 & 0.2240 & 0.8288 & -5.6425 & -2.1949 & 0.2937 & 0.5647 & 0.2161 & 0.0629 \\
  \hline
  MIMO-BEV & 0.3432 & 0.3991 & \textbf{1.0442} & 0.1852 & 0.2262 & \textbf{0.8073} & -5.3963 & \textbf{-2.5290} & 0.3589 & \textbf{0.5368} & \textbf{0.2004} & 0.0639 \\
 \Xhline{2\arrayrulewidth}
\end{tabular}
\end{table*}

The uncertainty results are laid out in \cref{tab:kitti_pp_ue} and \cref{tab:kitti_second_ue}. Scores are averaged over IoU thresholds from 0.5 to 0.95, with a 0.05 increment. For each threshold we follow Harakeh and Waslander~\cite{harakeh2021estimating} by calculating a score threshold that maximizes the F1 score in order to remove low scoring predictions. Temperature scaling~\cite{guo2017calibration} is also performed per class at each score threshold to calibrate the predictions.

MC dropout often outperforms the other models for classification NLL and Brier Score in the TP and FP\textsubscript{ML}, while MIMO-BEV performs well for FP\textsubscript{BG}s. The MC dropout model has predictions with higher confidence leading to lower scores. The MIMO-BEV model outputs lower confidence predictions, causing a better score for FP\textsubscript{BG}s.

For the regression scores, the MIMO-BEV model performs well on the KITTI dataset, with some best scores being taken by MC dropout. Once again the scores are low and are similar. The calibration errors are similar among all the models. The MCE values tend to be lowest for MIMO-BEV and ensemble, while the CE is lowest for MC dropout and MIMO-BEV.

\begin{figure}[t]
\centering
\includegraphics[width=1.0\columnwidth]{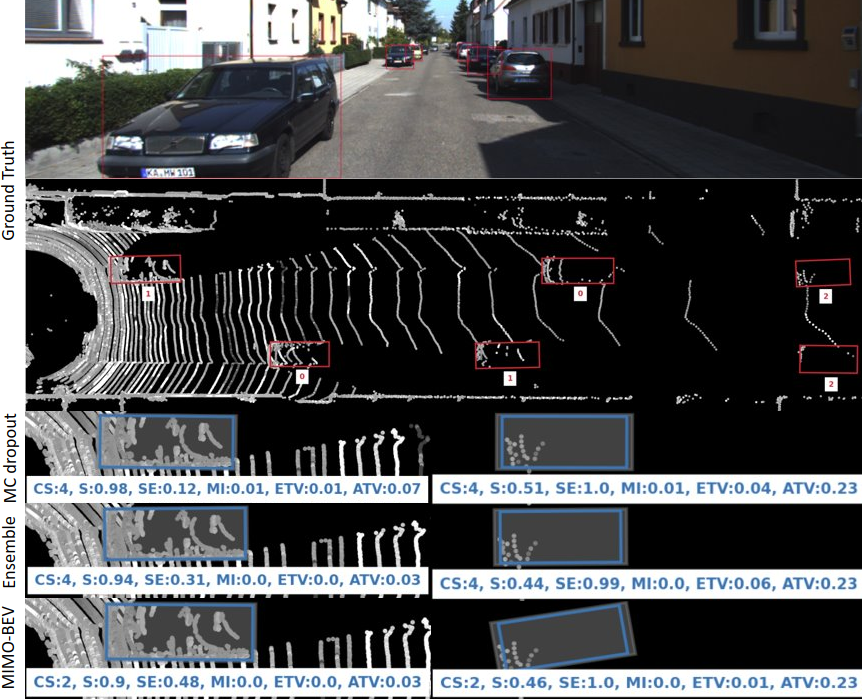}
\caption{Top: Ground truth objects are displayed for this KITTI frame on the camera image as well as the BEV projection along with the difficulty number. Bottom: Each row is a multi-output model with two TP predictions. The left one is the truncated car in the camera view and the right one is the occluded yellow car in the distance. The blue box is the mean cluster prediction while the grey box is the 95\% confidence interval for the width and height of the prediction.  Details on cluster size, softmax score, Shannon entropy, epistemic total variance and aleatoric total variance are included at the bottom.}
\label{f:bev_comparison}
\end{figure}

\Cref{f:bev_comparison} visualizes qualitative results for the PointPillars models trained on the KITTI dataset. The close TP prediction has a higher softmax score, lower SE, and lower total variances. The further TP prediction has a lower softmax score, and increased SE as well as ATV. The MI is zero or close to zero for each prediction, while the ETV is also low but increases with distance. ETV is the lowest for the MIMO-BEV model. Our results agree with previous work in this area displaying a large increase in aleatoric variance with distance away from the LiDAR sensor.

\section{CONCLUSION}
LiDAR-MIMO offers fast and high-quality detection and uncertainty estimation, which is crucial for safety-critical and resource-constrained robotic vision, such as found in autonomous driving. In our experiments, LiDAR-MIMO achieves consistently higher mAP than MC dropout, while approaching ensembles. Further, similar to the other two methods, LiDAR-MIMO produces high-quality uncertainty estimates as measured by calibration errors and scoring rules, but without the large runtime overhead of the other methods.
Our design for adapting MIMO to LiDAR-based 3D object detection relies on the composition of BEV feature maps, which is applicable to efficient 3D object detectors, such as PointPillars and SECOND. The stacking approach is likely to work well for other LiDAR-based 3D object detectors that are based on LiDAR BEV projections, but also LiDAR range images. Further, MIMO-ID can be used for any LiDAR-based detector, but it comes with increased data processing time. In general, MIMO is limited by the capacity of the underlying detector, which in our experiments allowed us to fit only two sub-networks.
In future work, we plan to explore LiDAR-MIMO with larger capacity networks and more detection heads, and adapting LiDAR-MIMO to other 3D detector designs, such as ones based on LiDAR range images.

%% file: sections/acknowledgement.tex
Several members within our lab also helped with the creation of this paper. Our co-op students Charles Zhang and Martin Ma contributed to the creation and testing of the calibration error calculations. The MASc student Spencer Delcore contributed to the uncertainty evaluation pipeline and validated many of the calculations that we use for the various metrics within the paper.